\documentclass[conference]{IEEEtran}
\IEEEoverridecommandlockouts
\usepackage{cite}
\usepackage{amsmath,amssymb,amsfonts}
\usepackage{algorithmic}
\usepackage{graphicx}
\usepackage{textcomp}
\usepackage{cuted}
\usepackage{xcolor}
\usepackage{amsmath, amsfonts}
\usepackage{breqn}
\usepackage{comment}
\usepackage{stfloats}
\usepackage{changepage}
\def\BibTeX{{\rm B\kern-.05em{\sc i\kern-.025em b}\kern-.08em
    T\kern-.1667em\lower.7ex\hbox{E}\kern-.125emX}}
\begin{document}

\title{Single-Entity Spiking Neuron Models: Survey}

\author{\IEEEauthorblockN{Leon Parepko}
\IEEEauthorblockA{\textit{Innopolis University}\\
Innopolis, Russia \\
leonparerpko@gmail.com}
\and
\IEEEauthorblockN{Danila Shulepin}
\IEEEauthorblockA{\textit{Innopolis University} \\
Innopolis, Russia \\
dshulepin2013@gmail.com}
\and
\IEEEauthorblockN{Albert Nasybullin}
\IEEEauthorblockA{\textit{Innopolis University}\\
Innopolis, Russia \\
levshaazz@gmail.com}

}

\maketitle

\begin{abstract}
In this work, we reviewed different approaches in mathematical modeling of biologically plausible neural systems. Models are characterized and classified based on their common features and special use cases. In addition to spiking models, different types of discrete and continuous analogs are considered to accurately simulate biological processes, including membrane potential dynamics. The models under investigation include neurons and various components encountered in neural systems and affected the dynamics. The selection of specific approaches was driven by their prevalence and innovative perspectives in order to enhance the relevance of the presented information.
\end{abstract} 

\section{Introduction}
The study of the brain functioning remains a complex problem. While the first and second generations of artificial neural networks (ANNs), based on McCulloch-Pitts model \cite{b1} with different activation functions, could not provide the accurate simulation of all the concepts, including the time-dependent voltage dynamics, underlie biological neural networks \cite{b2}. Large-scale brain models have emerged as a potential solution. Among these models, spiking neural networks (the third generation of ANNs) have shown promising results \cite{b2}. However, the challenge lies in accurately distinguishing between these models and selecting the most appropriate one, given the current lack of robust classification methods. 

In stark contrast to similar works [2,16], the present paper aims to explicate the underlying principles of various approaches and provide a multifaceted survey that includes a models features comparison and an analysis of the correlated problems that these ones solve.

In order to achieve this goal, we have categorized all neural systems models into two major sets: single-entity models, which are also referred to as single compartmental models; and composite models, or multi-compartmental models \cite{b3}. Nonetheless, we shall focus solely on the former class in this paper while providing a comprehensive classification of both categories in future work.

In the context of single-entity models, a single computational unit initially represents the cell, which is described by a mathematical model. It ought to be noted that such models may comprise multiple equations. The key distinction here is that cell is not constructively divided into separate components (e.g., axon, dendrites, synapses), as is done in multi-compartmental modeling, where each component corresponds to a distinct mathematical model.

\section{Integrate and Fire models} \label{IF}
The Integrate and Fire (IF) [4,5] is a collection of biophysically meaningless models, characterized by ordinary differential equations (ODEs). The ones rely on two fundamental concepts: integration - summation of input signals, and firing - emitting a spike beyond a particular threshold $V_{th}$. These models are low-dimensional and neglect ion-channel dynamics. Corresponding abstraction could be constructed by an electrical circuit, where the capacitor represents the passive membrane $V$. The basic IF model (1) aggregates the input current $I_{ext}$ and generates a spike as described above.

\begin{dmath}
    \begin{cases}
    C\frac{dV}{dt}=I_{ext}\\
    \textrm{if  }V\geq V_{th} \textrm{, then }V=V_{reset}
    \end{cases}
\end{dmath}

Alternatively, the more practical leaky-IF (LIF) model (2), which involves a constant membrane leak through a resistor, is commonly used. 

\begin{dmath}
    \begin{cases}
    \frac{dV}{dt}=I_{ext}-g(V-V_0)\\
    \textrm{if  }V\geq V_{th} \textrm{, then }V=V_{reset}
    \end{cases}
\end{dmath}

The membrane capacity is denoted by $C$, $V_0$ represents the resting potential, and $g$ signifies the leak conductance.

LIF modifications can introduce additional functionalities, such as refractory period, which could be implemented by forcing neuron to ignore the after-spike input current for some time $\tau_{ref}$ \cite{b6}. As another example, adaptive LIF can be implemented in various ways, which relies on an additional ODE to describe adaptation dynamics, including the method in \cite{b7}. In equation (3), $V_{th}$ increases by $d$ after each emitted spike and has a distinct leak, denoted by $a$. 

\begin{dmath}
    \begin{cases}
    \frac{dV}{dt}=I_{ext}-g(V-V_0)\\
    \frac{dV_{th}}{dt}=a(V-V_{th})\\    
    \textrm{if  }V\geq V_{th} \textrm{, then }\begin{cases}
        V =V_{reset}\\
        V_{th} =V_{th}+d 
        \end{cases}
    \end{cases}
\end{dmath}

One more popular LIF modification is the quadratic LIF \cite{b8} (4), which, unlike basic LIF, can exhibit bistable states and can alter activity type in response to varying input stimuli. Neurons with these properties, known as meta-dynamic neurons (MDNs), can process spatial and temporal information concurrently, leading to better network generalization. However, experiments on another LIF-based neuron model with sigmoid input activation, namely the first-order MDN, indicate poor performance on temporal tasks relative to spatial ones \cite{b9}. U. Chialva et al. \cite{b10} presented an overview of IF model construction, while  [11,12] explore a variety of IF models.

\begin{dmath}
    \begin{cases}
    \frac{dV}{dt}=I_{ext}-a(V-V_0)(V-V_{th})\\
    \textrm{if  }V\geq V_{th} \textrm{, then }V =V_{reset}
    \end{cases}
\end{dmath}

IF-based neurons present a major advantage in their simplicity, which enables their effortless implementation via electrical circuits in neuromorphic processors  \cite{b13}. However, IF neurons fall short in full representation of activity in biological neurons. For instance, they fail to account for membrane potential dynamics beyond the threshold value and emit binary spikes, omitting the characteristics real action potentials display, such as duration, amplitude, and shape. Hence, the foregoing models are not recommended for biological simulations [4,5].

\section{Hodjkin-Huxley based models}
The Hodjkin-Huxley (HH) model  \cite{b14}, a biophysically plausible neural model widely employed in neuroscience, comprises four ODEs. The model characterizes the neuron as the aggregation of ion channel currents which, in turn, results in a change of membrane potential $V$ over time $t$. These equations (5) were derived using a probabilistic kinetic and Markov modeling  \cite{b15} of the sodium and potassium ion channels, as well as the pump.

\begin{dmath}
    \begin{cases}
    \scalebox{0.74}[1.0]{\( C\frac{dV}{dt}=I_{ext}-\overline{g}_Kn^4(V-V_K)-\overline{g}_{Na}m^3h(V-V_{Na})-\overline{g}_L(V-V_L) \)}\\
    \frac{dx}{dt}=a_x(V)(1-x)-b_x(V)x
    \end{cases}
\end{dmath}

The behavior of ion channels is governed by the activation and inactivation variables $m, n, h$ standing for sodium and potassium. Notably, the ion pump is exempt from these variables due to its static nature. The maximum channel conductance is denoted by $\overline{g}$. Specifically, the channel dynamics is modeled by three equations using $x$ to describe $m$, $n$, and $h$. Although similar in form, these equations differ in their values of $a_x$ and $b_x$, which have been determined empirically.

The HH model, with its flexibility and high accuracy in describing neural properties and dynamics  [2,14,16], has become a common approach in neural modeling. Nevertheless, it suffers from certain limitations. In contrast to biological neurons, the activation and inactivation in the HH model are kinetically independent  \cite{b15}. Furthermore, the model is analytically complex and computationally demanding, leading to low performance in large-scale simulations. Accordingly, researchers have sought to develop HH-based models that simplify the HH model. U. Chialva et al.  \cite{b10} classified the ones into construction types: variable reduction, which involves the systematic reduction of state variables; no spiking currents, which consist solely of ionic currents (e.g., $I_{Na}, I_{K}$) without any spiking ones; phenomenological models, which rely on simplifications of phase planes to mimic real neurons, thereby capturing essential patterns; linearized models, which involve linearization of the model; and integrate and fire models, a specific type of model that employs the passive membrane equation (described in section \ref{IF}).

To clarify, a number of popular HH-based models have been enumerated below. The FitzHugh-Nagumo model, a two-dimensional continuous phenomenological simplification of the HH model, was initially developed by FitzHugh and later generalized by Nagumo  \cite{b17}. This model is based on separating ion currents according to their time scales, by assuming that certain ion channels undergo rapid activation and achieve stability quite swiftly. The sum of such channels' currents can then be approximated as $f(V)$ (in this case $\frac{u^3}{3}$).

\begin{dmath}
    \begin{cases}
    \frac{dV}{dt}=\varepsilon(u-a-bV)\\
    \frac{du}{dt}=u-\frac{u^3}{3}-V+I_{ext}
    \end{cases}
\end{dmath}

The variable $u$ describes relatively slow ion currents that affect the recovery of resting membrane potential. This model is capable of operating in three modes, including single spike generation, periodic spiking, and subthreshold fluctuations. However, the channel conductance parameters $a$ and $b$ can be difficult to calibrate precisely, thus leading to the development of a range of novel variations designed to simulate diverse types of spiking activities. Such variations include models with non-linear recovery mechanisms  \cite{b18}, adaptive controlled  \cite{b19}, and even discrete analog as the Courbage-Nekorkin model  \cite{b20}.

Izhikevich neuron (IZ) model  \cite{b21} is highly promising for large-scale brain simulations, particularly in the cortical area, due to its computational efficiency. This model comprises two ODEs, which were derived from bifurcation analyses and phase plane simplification of the HH model. The constants used were obtained by fitting the dynamics of cortical neurons and can be further adjusted to match other types of neurons.

\begin{dmath}
    \begin{cases}
    \frac{dV}{dt}=0.04V^2+5V+140-u+I_{ext}\\
    \frac{du}{dt}=a(bV-u)\\
    \textrm{if  } V \geq30\textrm{mV, then}\begin{cases}
        V \leftarrow c \\
        u \leftarrow u+d 
        \end{cases}
    \end{cases}
\end{dmath}

Slow recovery variable $u$ incorporates the currents from sodium and potassium ion channels, taking into account their activation and inactivation periods. The recovery variable time scale and sensitivity parameters are represented by $a$ and $b$, while the parameters $c$ and $d$ determine the after-spike reset value for the membrane potential and the recovery variable, respectively. One noteworthy observation is that if $a$ is notably small while $c$ is relatively large, the model behaves similarly to an IF model by discarding the recovery variable. 

Building on primary IZ model, second-order MDNs were created, which displayed a substantial improvement in network accuracy for spatio-temporal tasks  \cite{b9}. Besides, the IZ neuron can be discretized with a 1ms resolution  \cite{b16} and may facilitate simulating of tens of thousands of discrete neurons with concerningly small computational resources and resulting in sufficiently accurate spiking activity.

The utility of the IZ neuron model is particularly evident for tasks where biophysically meaningless models are suitable, although it is not recommended for research on the ion-channel level of signal transmission  \cite{b2}. It is frequently used in spiking neural networks (SNNs)  \cite{b22} as well as in benchmarking.

\section{Dendritic Neuron Models}
Traditionally, in the McCulloch-Pitts model, the function of the dendrites was represented as a weighted sum of the synaptic inputs. However, recent researches show that dendrites play a more active role in the operation and local information processing. The dendritic neuron model (DNM) takes into account the shape of the dendrite and the interaction between synapses, which is more accurate than the McCulloch-Pitts model  \cite{b27}. The DNM  \cite{b28} consists of four layers: synaptic, dendritic, membrane and cell.

The synaptic layer applies equation (8) to process input signals and transmits them to the dendritic layer.

\begin{dmath}
    S_{i,m} = \sigma(k(w_{i,m}x_{i} - q_{i,m}))
\end{dmath}

 Here $x_{i}$ is the $i^{th}$ input feature, $w_{i,m}$ and $q_{i,m}$ are the weight and bias of the synapses on $m^{th}$ dendritic branch for $i^{th}$ feature, $k$ is a dimensionless hyperparameter and $\sigma$ is a sigmoid function.

The dendritic layer multiplies the outputs of the synaptic layer on a respective branch and passes them to the membrane layer, which sums the results of each dendritic branch (9). 

The cell layer receives the potential from the membrane layer. As soon as it exceeds the threshold, the neuron generates signals and transmits them to other neurons via the axon, which is described using a sigmoid function (9).

\begin{dmath}
    O = \sigma(k(\sum_{m = 1}^{M}\prod_{i = 1}^{I}S_{i,m} - \theta))
\end{dmath}

 Here $I$ and $M$ are the numbers of inputs and dendritic branches correspondingly, also $\theta$ is the threshold value.

The DNM output can fall into one of four connection cases, depending on the values of parameters $q$ and $w$. These cases include direct ($0 < q_{i;m} < w_{i;m}$); inverse ($w_{i;m} < q_{i;m} < 0$); constant 1 ($q_{i;m} < 0 < w_{i;m}$) and constant 0 connections ($0 < w_{i;m} < q_{i;m}$). Based on the connection cases, DNM discards unnecessary synapses and dendritic branches to create a unique model structure for specific tasks via synaptic or dendritic pruning.

Compared to the models above, DNM does not describe the dynamic aspects of the systems. Also, its main limitation is its suffering from the curse of dimensionality  \cite{b28}. In addition, the simplified structure can be easily implemented by comparators and logic gates AND, OR, and NOT, without affecting the accuracy of the model  \cite{b28}. One of the most promising applications of DNM is the analysis of mechanisms and principles for the operation of synapses and dendrites in biological neurons.

\section{Artificial Neural Networks}
The second generation of ANNs lacks biological plausibility and yet can discretely simulate complex single-neuron behavior. Notably, among different model architectures examined, convolutional-long short-term memory (CNN-LSTM) showed the best performance in simulation  \cite{b29}. After training and validating ANN on a single-compartmental neuron model based on  \cite{b30} realization (including inhibitory and excitatory synaptic currents), comparative simulations were conducted on the primary test such as threshold dynamics, and signal summation, among others. This model can replicate the membrane potential, ion channel conductances (sodium and potassium), AMPA-NMDA receptors containing and inhibitory synapse non-linearities, and accurately simulate multi-compartmental information from dendritic trees.

Notwithstanding, the CNN-LSTM architecture tends to ignore long-duration firing patterns  \cite{b21} due to the temporary property of model memory. Meanwhile, an additional custom layer of IZ neurons solves this issue efficiently with only five more trainable ANN parameters compared to the millions of initial model parameters.

The primary feature of this approach is its versatility, as after training it can mimic any kind of activity through generalization (i.e., neuron populations, groups) or specialization (i.e., synapses, dendrites). It could be applied in simulations of atypical neural behavior that traditional approaches cannot handle, particularly with regard to neural disease simulations. Large populations of neurons can be numerically computed quickly and efficiently using hardware accelerators such as GPUs. In contrast to ODE approach, where each cell is represented by a set of equations, ANNs utilize a single computational graph for multiple cells, enabling easy parallelization. Generalization properties of ANNs could be used in model dimensionality reduction, allowing smaller neuron populations to be generalized into larger single ones represented by ANN. Despite these advantages, similar to any phenomenological model, ANNs can not solve out-of-domain data problem  \cite{b31}. Also, the time-cost of training prior to simulation commencement can negate the advantages of using ANNs.

\section{Conclusion}
Our study has provided an overview of fundamental approaches for modeling single-entity neural systems, with a particular emphasis on biologically-plausible models. Models' categorization was described in terms of their benefits and limitations, while also highlighting the relevance and advantages of novel approaches in comparison to traditional ones. This paper may serve as a guideline for the selection of appropriate modeling approaches. The outcomes of our research are reflected in Table \ref{table:modelsComparison} provides a comprehensive comparison of even more models beyond those discussed, using relevant and objective parameters. Important to note that the results presented in the "Number of Parameters" and "Number of FLOPs" columns may vary in different model realizations. Thus, $C$ and $I_{ext}$ are excluded while counting the parameters to minimize possible model inconsistencies, and each primary mathematical operation (including relation, division, power, exponentiation, etc.) is considered as a single floating point operation.

\begin{table*}[htp]
\vspace{-0.2cm}
\caption{Comparison of Single-entity Neurons}
\begin{center}
\begin{adjustwidth}{0.25cm}{}
\begin{tabular}{|c|c|c|c|c|c|c|c|}
\hline
\textbf{}&\multicolumn{7}{|c|}{\textbf{Comparative characteristics}} \\
\cline{2-8} 
\textbf{Neural Models} & \textbf{\textit{Discrete or}}& \textbf{\textit{HH-based}} & \textbf{\textit{Number of}} & \textbf{\textit{Number of}} & \textbf{\textit{Number of}} & \textbf{\textit{Experimental}} & \textbf{\textit{Izhikevich}} \\
\textbf{} & \textbf{\textit{Continuous}} & \textbf{\textit{}} & \textbf{\textit{Variables}} & \textbf{\textit{Parameters}} & \textbf{\textit{FLOPs}} & \textbf{\textit{Evidence}} & \textbf{\textit{Dynamics}}  \\
\hline
HH & Continuous & \checkmark & 4 & 11 & 1200 & \checkmark & Any \\
\hline
Izhikevich & Continuous & P & 2 & 5 & 13 & \checkmark & Any \\
\hline
IF & Continuous & IFB & 1 & 3 & 1 & \texttimes & FS \\
\hline
LIF & Continuous & IFB & 1 & 4 & 5 & \texttimes & RS, FS \\
\hline
Adaptive IF (active threshold) & Continuous & IFB & 2 & 5 & 10 & \texttimes & RS, FS, LTS \\
\hline
Exponential IF & Continuous & IFB & 1 & 5 & 11 & \texttimes & RS, FS \\
\hline
Quadratic IF & Continuous & IFB & 1 & 4 & 7 & \texttimes & RS, FS \\
\hline
IF Or-Burst & Continuous & IFB & 2 & 8 & 26 & \checkmark & RS, IB, C, FS, TB \\
\hline
Morris-Lecar & Continuous & WSC & 3 & 12 & 600 & \checkmark & RS, FS \\
\hline
BVP & Continuous & WSC & 2 & 4 & 72 & \textendash & RS, C, FS, LTS, R \\
\hline
FHN & Continuous & P & 2 & 6 & 72 & \textendash & RS, FS, LTS, R \\
\hline
Wilson & Continuous & P & 2 & 6 & 180 & \checkmark & Any \\
\hline
Hindmarsh-Rose & Continuous & P & 3 & 7 & 120 & \textendash & RS, IB, LTS, TB \\
\hline
MDN (1st order) & Continuous & IFB & 1 & $I + 3$ & $I \cdot 2 + 7$ & \texttimes & RS, FS \\
\hline
MDN (2nd order) & Continuous & P & 2 & $I + 5$ & $I\cdot 2+13$ & \checkmark & Any \\
\hline
CNN-LSTM+IZ & Discrete & \texttimes & Any & $1.95 \cdot 10^6 + 5$ & \texttimes & \checkmark & Any \\
\hline
CNN-LSTM & Discrete & \texttimes & Any & $1.95 \cdot 10^6$ & \texttimes & \checkmark & RS, IB, C, FS, TB \\
\hline
DNM & Discrete & \texttimes & $I$ & $2 \cdot I \cdot M + 2$ & $6 \cdot I \cdot M + 5$ & \textendash & \textendash \\
\hline
Spike-Response & Discrete & IFB & 2 & 4 & 50 & \textendash & Any \\
\hline
Rulkov model & Discrete & \texttimes & 2 & 3 & 8 & \textendash & RS, IB, C, FS, TB \\
\hline
Courbage–Nekorkin model & Discrete & P & 2 & 4 & 13 & \textendash & \textendash \\
\hline
Chialvo model & Discrete & \texttimes & 2 & 4 & 9 & \textendash & RS, IB, C, FS, TB \\
\hline

\multicolumn{1}{l}{\vspace{-0.2cm}}\\

\multicolumn{1}{l}{\hspace*{-0.1cm}\checkmark: Matches the property;} & \multicolumn{3}{l}{$^{\mathrm{\phantom{*}}}$WSC: Without Spiking Currents;} & \multicolumn{3}{l}{$^{\mathrm{\phantom{*}}}$\hspace*{-0.8cm}RS: Regular Spiking; TB: Tonic Burst;} & \multicolumn{1}{l}{\hspace*{-1.5cm}$I$: number of inputs;}\\

\multicolumn{1}{l}{\hspace*{-0.1cm}\texttimes: Doesn't match the property;} & \multicolumn{3}{l}{$^{\mathrm{\phantom{*}}}$P: Phenomenological;} & \multicolumn{3}{l}{$^{\mathrm{\phantom{*}}}$\hspace*{-0.8cm}C: Chattering; FS: Fast Spiking;} & \multicolumn{1}{l}{\hspace*{-1.5cm}$M$: number of dendritic branches.}\\

\multicolumn{1}{l}{\hspace*{-0.1cm}\textendash: No available information;} & \multicolumn{3}{l}{$^{\mathrm{\phantom{*}}}$IFB: IF-based.} & \multicolumn{3}{l}{$^{\mathrm{\phantom{*}}}$\hspace*{-0.8cm}R: Resonator; IB: Intrinsically Bursting;}\\

\multicolumn{1}{l}{\hspace*{-0.1cm}Any: All of the above.} & \multicolumn{3}{l}{$^{\mathrm{\phantom{*}}}$} & \multicolumn{3}{l}{$^{\mathrm{\phantom{*}}}$\hspace*{-0.8cm}LTS: Low-threshold Spiking.}\\

\end{tabular}
\label{table:modelsComparison}
\end{adjustwidth}
\end{center}
\end{table*}

\end{document}